\documentclass[sigconf]{acmart}

\renewcommand\footnotetextcopyrightpermission[1]{} 
\usepackage{amsmath}
\usepackage{multirow}
\usepackage{booktabs}
\usepackage{xcolor}
\usepackage{url}
\usepackage{graphicx}
\usepackage{subcaption}
\usepackage{comment}
\usepackage{caption}
\usepackage{todonotes}

\copyrightyear{2025}

\title{Large Scalable Cross-Domain Graph Neural Networks for Personalized Notification at LinkedIn
}

\author{Shihai He}
\authornote{Corresponding author.}
\affiliation{
  \institution{LinkedIn}
  \city{Sunnyvale}
  \state{CA}
  \country{USA}
}
\email{she1@linkedin.com}

\author{Julie Choi}
\authornotemark[1]
\affiliation{
  \institution{LinkedIn}
  \city{Sunnyvale}
  \state{CA}
  \country{USA}
}
\email{julchoi@linkedin.com}

\author{Tianqi Li}
\affiliation{
  \institution{LinkedIn}
  \city{Sunnyvale}
  \state{CA}
  \country{USA}
}
\email{tiali@linkedin.com}

\author{Zhiwei Ding}
\affiliation{
  \institution{LinkedIn}
  \city{Sunnyvale}
  \state{CA}
  \country{USA}
}
\email{zding@linkedin.com}

\author{Peng Du}
\affiliation{
  \institution{LinkedIn}
  \city{Sunnyvale}
  \state{CA}
  \country{USA}
}
\email{pedu@linkedin.com}

\author{Priya Bannur}
\affiliation{
  \institution{LinkedIn}
  \city{Sunnyvale}
  \state{CA}
  \country{USA}
}
\email{pbannur@linkedin.com}

\author{Franco Liang}
\authornote{Work done while at LinkedIn.}
\affiliation{
  \institution{University of California, Davis}
  \city{Davis}
  \state{CA}
  \country{USA}
}
\email{frnliang@ucdavis.edu}

\author{Fedor Borisyuk}
\affiliation{
  \institution{LinkedIn}
  \city{Sunnyvale}
  \state{CA}
  \country{USA}
}
\email{fborisyuk@linkedin.com}

\author{Padmini Jaikumar}
\affiliation{
  \institution{LinkedIn}
  \city{Sunnyvale}
  \state{CA}
  \country{USA}
}
\email{pjaikumar@linkedin.com}

\author{Xiaobing Xue}
\affiliation{
  \institution{LinkedIn}
  \city{Sunnyvale}
  \state{CA}
  \country{USA}
}
\email{xxue@linkedin.com}

\author{Viral Gupta}
\affiliation{
  \institution{LinkedIn}
  \city{Sunnyvale}
  \state{CA}
  \country{USA}
}
\email{vigupta@linkedin.com}

\begin{document}

\begin{abstract}

Notification recommendation systems are critical to driving user engagement on professional platforms like LinkedIn. Designing such systems involves integrating heterogeneous signals across domains, capturing temporal dynamics, and optimizing for multiple, often competing, objectives. Graph Neural Networks (GNNs) provide a powerful framework for modeling complex interactions in such environments.

In this paper, we present a cross-domain GNN-based system deployed at LinkedIn that unifies user, content, and activity signals into a single, large-scale graph. By training on this cross-domain structure, our model significantly outperforms single-domain baselines on key tasks, including click-through rate (CTR) prediction and professional engagement. We introduce architectural innovations including temporal modeling and multi-task learning, which further enhance performance.

Deployed in LinkedIn’s notification system, our approach led to a 0.10\% lift in weekly active users and a 0.62\% improvement in CTR. We detail our graph construction process, model design, training pipeline, and both offline and online evaluations. Our work demonstrates the scalability and effectiveness of cross-domain GNNs in real-world, high-impact applications.

\end{abstract}

\maketitle


\section{Introduction}
\label{sec:intro}
%

Modern professional networks, such as LinkedIn, expose users to a diverse array of content domains, including connections' feed posts, out-of-network content, job recommendations, and trending news, with the goal of improving user engagement, facilitating opportunity discovery, and strengthening professional relationships. While each content domain typically has a dedicated surface for user interaction, the notification system serves as a unified channel to inform users of recent updates across these domains and to stimulate further engagement. Integrating the signals from all these domains will help the notification system to precisely push the right content to the right user.

However, designing an effective notification system presents several distinctive challenges:
\begin{itemize}
\item \emph{Underutilization of cross-domain signals}: An effective system must integrate heterogeneous signals originating from multiple content domains, which are often fragmented and underexploited.
\item \emph{Temporal sensitivity}: Notifications must be generated and delivered in a timely manner, as their utility and impact are highly dependent on recency relative to user activity.
\item \emph{Multi-objective optimization}: The system must simultaneously optimize for multiple, and sometimes competing, objectives, such as immediate user interaction (e.g., click-through rate), long-term engagement, and overall user satisfaction.
\end{itemize}
Since their introduction ~\cite{gori2005new, scarselli2008graph, wu2020comprehensive}, Graph Neural Networks (GNNs) have been widely explored in the context of social networks ~\cite{fan2019graph, sharma2024survey, li2023survey} and have seen increasing adoption in applications of the IT industry ~\cite{zhao2025gigl, ying2018graph, zhang2023graphstorm, LiGNN2024KDD, liu2024linksage}. Their ability to effectively model complex relationships and propagate information across graph structures makes them particularly well-suited for incorporating cross-domain signals. In the context of notification systems, GNNs offer a powerful mechanism to learn rich and context-aware representations of both users and content items, thereby enabling more relevant and personalized notifications.
\newline
In this paper, we present a \emph{Cross-domain Graph Neural Network} that merges signals from multiple LinkedIn domains into a \emph{single} heterogeneous graph and utilizes temporal modeling and multitask learning in the GNN to address the above-mentioned challenges. This work builds upon our earlier successes in GNN productionization with LiGNN~\cite{LiGNN2024KDD}. While LiGNN primarily focused on individual domains like Jobs and Feed, cross-domain GNN extends this approach to encompass Notifications, Feed, Email, and other surfaces, thereby leveraging the complete relational structure at LinkedIn.

\smallskip

\noindent\textbf{Contributions.} Our key contributions include:
\begin{itemize}
\item \textbf{Unified Graph Construction:} We detail a systematic approach to harmonizing data from multiple domains into a large-scale heterogeneous graph (15TB) with 8.6 billion nodes and over 100 billion edges, capturing user-content interactions and social connections across LinkedIn's ecosystem.

\item \textbf{Temporal Modeling:} We introduce a time-aware component that captures the temporal dynamics of edges, significantly improving predictive accuracy in rapidly changing domains like notifications where recency strongly influences relevance.

\item \textbf{Multi-task Learning Architecture:} We adapt Multi-gate Mixture-of-Experts (MMoE) for graph-based prediction tasks, allowing the model to learn both task-specific and shared knowledge across different objectives (e.g., notification clicks and professional interactions).

\item \textbf{Production System Design:} We describe our scalable pipeline for training and serving the cross-domain GNN on the LinkedIn scale, supporting real-time inference for hundreds of millions of members while maintaining low latency.

\item \textbf{Empirical Validation:} We present comprehensive offline and online A/B test results demonstrating that the Cross-domain GNN outperforms domain-specific baselines, with particularly significant gains in notification personalization (+0.62\% CTR, +0.10\% WAU).

\end{itemize}

The remainder of this paper is structured as follows. Section~\ref{sec:related} provides an overview of related work in large-scale GNNs and cross-domain learning. Section~\ref{sec:unified-graph} explains our approach to building a unified and heterogeneous graph on LinkedIn. Section~\ref{sec:model} introduces our GNN architecture, including temporal extensions and multi-task learning (MTL). Section~\ref{sec:production} outlines the production pipeline and deployment details. We present experimental results in Section~\ref{sec:experiments}, discuss broader implications in Section~\ref{sec:discussion}, and conclude in Section~\ref{sec:conclusion}.

\section{Related Work}
\label{sec:related}

\subsection{Graph Neural Networks in Industrial Systems}

Graph Neural Networks (GNNs) have become widely used in industrial applications such as recommendation systems and knowledge graphs, building on architectures like GCN~\cite{kipf2017semi}, GraphSAGE~\cite{hamilton2017inductive}, GAT~\cite{velivckovic2018graph}, and PPRGAT~\cite{choi2022pprgat}. Early large-scale deployments such as PinSAGE~\cite{ying2018graph} demonstrated how to scale GNNs for recommendation at Pinterest.

At LinkedIn, LiGNN~\cite{LiGNN2024KDD} marked a key step in operationalizing GNNs, initially for single-domain use cases. Our work extends this to a cross-domain setting, addressing the challenges of modeling heterogeneous interactions across multiple product surfaces.




 \subsection{Cross-Domain Learning}

Recommendation systems often suffer from data sparsity~\cite{guo2017} and cold-start issues~\cite{zhang2019}. Cross-Domain Recommendation ~\cite{conet2018,minet2020,hetergraph2024,cui2020,ning2023} mitigates these problems by aggregating user data from multiple domains. Many Cross-Domain Recommendation methods use separate models per domain and leverage transfer~\cite{conet2018,minet2020} or simultaneous learning~\cite{cui2020,ning2023}, limiting their ability to learn holistic user preferences.

In contrast, we model all domains in a unified industrial-scale graph. This allows message passing to naturally propagate cross-domain signals during embedding generation, enabling global preference learning.



\subsection{Temporal Graph Modeling}

Temporal graphs represent evolving structures where edges, nodes, or attributes change over time~\cite{skarding2021foundations, han2021dynamic, longa2023graph, feng2024comprehensive}. Models like TGAT~\cite{xu2020inductive} and TGN~\cite{rossi2020temporal} integrate temporal dynamics into message passing but often face complexity and scalability challenges.

Recent methods~\cite{liu2024tgtod, min2022transformer, LiGNN2024KDD, beladev2023graphert} use transformer-based models to encode temporal patterns more efficiently. We build on the temporal modeling framework in LiGNN~\cite{LiGNN2024KDD}, adapting it to the Notification system for time-sensitive personalization tasks.



\subsection{Multi‑Task Learning for Graph Neural Networks}
\label{sec:related_mtl_gnn}
Multi-task learning (MTL) has been widely adopted in various fields to enhance model performance and reduce maintenance overhead. Beyond early works that focused on hard or soft parameter sharing across task heads~\cite{caruana1997multitask, long2017learning, misra2016cross, ruder2019latent}, recent architectures like MMoE~\cite{ma2018modeling}, PLE~\cite{tang2020progressive}, and AdaTT~\cite{li2023adatt} incorporate gating mechanisms to dynamically balance shared and task-specific knowledge.  

In the context of graph data, early work demonstrated the benefits of jointly predicting node-level and graph-level labels, showing gains for both granularity levels~\cite{holtz2019multitask,xie2020repr}. 
A meta‑learning view later emerged: first learn task‑agnostic node embeddings, then attach lightweight heads for specific objectives, which speeds adaptation to new tasks~\cite{buffelli2021meta}. 
Our work extends the MMoE framework~\cite{ma2018modeling} to GraphSAGE~\cite{hamilton2017inductive}, enabling cross-domain GNN modeling for predicting multiple types of user–item interactions.



\section{Unified Graph Construction}
\label{sec:unified-graph}

In this section, we present a generalizable framework for constructing a unified, cross-domain heterogeneous graph centered on member engagement. Each domain-specific graph is designed with three types of edges: engagement, affinity, and intrinsic. Each node in the graphs is represented by a set of node features. By strategically merging selected domain graphs, we build a comprehensive cross-domain graph that integrates data across multiple LinkedIn product domains, including Email, Feed, and Notifications. This design is inherently scalable and can accommodate additional domains as needed.

The integration of data across domains into a unified graph could raise questions around privacy, data protection and regulatory alignment. Organizations might explore practices such as excluding data from individuals without explicit consent, narrowing data use to compatible purposes, and applying anonymization strategies to reduce identifiability. These steps may support efforts to align with evolving legal and ethical expectations.

\subsection{Edge Construction}

To effectively model user interactions and content relevance in our GNN, we capture the multifaceted relationships between entities through a structured hierarchy of edge types:

\begin{itemize}
\item \textbf{Engagement Edges:} These directed edges are fundamental for our notification modeling objectives, connecting members to the specific content items they engage with (e.g., ``member M1 clicked on notification N''). These edges provide direct supervision signals and anchor the graph in observable behavioral patterns. They carry valuable temporal information about when the interaction occurred.

\item \textbf{Affinity Edges:} To enrich representations and generalize beyond sparse engagement signals, we introduce bidirectional edges connecting viewers and creators of engaged content over a longer period For example, ``member M2 has engaged with feed posts created by member M1'' would be represented by a bidirectional edge between M1 and M2. These edges help our GNN propagate preferences and behavioral patterns through the latent social graph, inferring potential interests between members even without direct recent engagement.

\item \textbf{Intrinsic Edges:} These edges capture slowly evolving or static entity attributes, such as ``member M1 has title T'' or ``notification N mentioned feed post P.'' They encode identity and contextual attributes that ground nodes in semantically meaningful metadata, improving the GNN's ability to reason over both behavior and identity.

\end{itemize}

\begin{figure}[t]
\centering
\includegraphics[width=1.0\linewidth]{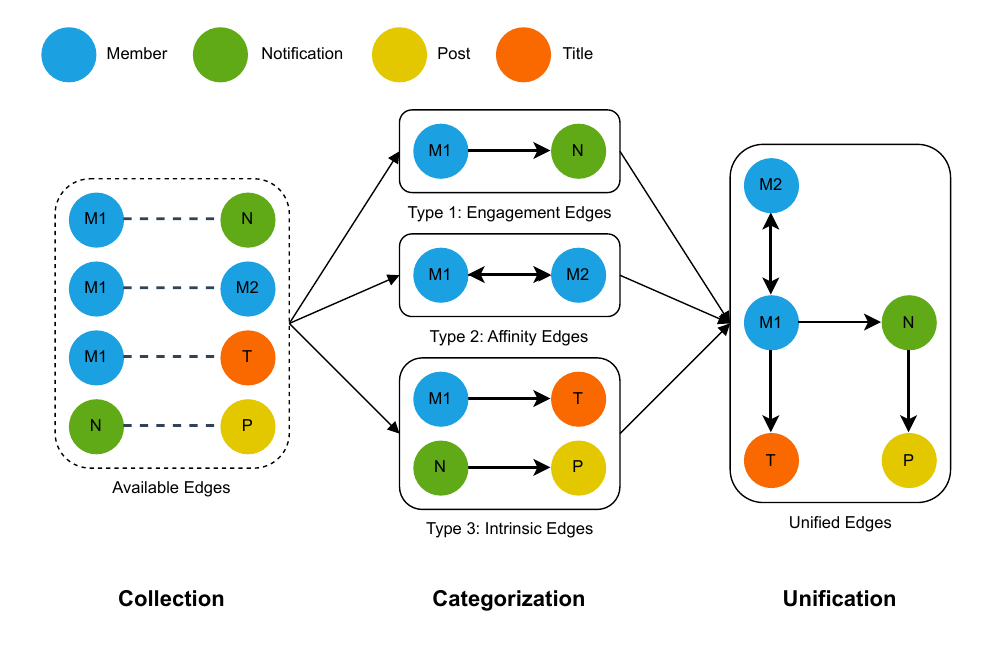}
\caption{Our generalizable framework to collect all available edges, categorize them into engagement, affinity, or intrinsic types, and unify all three types into a comprehensive cross-domain graph.}
\label{fig:unified-edges}
\end{figure}

By constructing these three types of edges, we create a graph structure that captures both immediate actions and the broader relational and semantic landscape in which those actions take place. This framework, illustrated in Figure~\ref{fig:unified-edges}, allows us to merge domain-specific graphs into one unified cross-domain graph, providing a holistic view of member engagements across the LinkedIn platform. The current unified graph contains several hundred billion edges, forming a rich substrate for cross-domain learning.

\subsection{Node Features}

Each node in our cross-domain graph is represented by a set of features that provide essential initial state information for GNN message passing. We incorporate two primary types of node features:

\begin{itemize}
\item \textbf{Dense Features:} These capture the semantic content associated with each node. Examples include BERT-generated text embeddings for content items (e.g., posts, notifications, emails) and aggregated content embeddings for members based on their historical engagement patterns.

\item \textbf{Categorical Features:} These encode discrete attributes of each node, such as industry, seniority, or content type, allowing the GNN to distinguish between different entity categories and learn type-specific patterns.

\end{itemize}

These node features provide the initial representation for each entity, helping the GNN identify nodes and learn useful patterns inductively through message passing across the graph structure.

\subsection{Temporal Graph Construction}

We transform our static graph into a temporal graph by assigning timestamps to the engagement edges, enabling us to capture dynamic changes over time. This temporal dimension serves two critical purposes:

\begin{enumerate}
\item It enables temporal sequence modeling within our GNN architecture, allowing us to extract member action patterns that respect chronological ordering.

\item It prevents information leakage during training by ensuring we only sample edges that occurred before the label timestamp when making predictions.

\end{enumerate}

\subsection{Data Partitioning}

To ensure rigorous evaluation and prevent data leakage, we partition our graph data into four distinct periods:

\begin{itemize}
\item \textbf{Graph Construction Period (35 days):} Used to build the unified graph structure with all necessary nodes and edges.

\item \textbf{Training Data Period (14 days):} Period from which labeled examples are drawn for model training.

\item \textbf{Validation Data Period (7 days):} Period from which independent examples are drawn to evaluate model performance.

\end{itemize}

This temporal partitioning ensures that our model is evaluated on future data, mirroring real-world deployment scenarios where predictions must be made on unseen future interactions.

\section{Cross-domain GNN Architecture}
\label{sec:model}

Our Cross-domain GNN employs a two-tower architecture with separate entity encoders for member and item (notification) representations, as illustrated in Figure~\ref{fig:gnn-arch}. Each entity encoder processes its respective input nodes through a feature transformation module (node encoder) followed by specialized aggregators. The member-side encoder incorporates both a heterogeneous message passing module (Sage aggregator) and a transformer based temporal aggregation module to capture dynamic interaction patterns, while the item-side encoder focuses on heterogeneous message passing to model contextual relationships around the content.
The member and item embeddings produced by the encoders are fed into a decoder to estimate the probability of a link between the pair.
We also support MTL for simultaneous prediction of multiple tasks, adopting the MMoE framework \cite{ma2018modeling}, which incorporates both shared and task-specific experts - each implemented as an entity encoder.

\begin{figure}[t]
\centering
\includegraphics[width=0.75\linewidth]{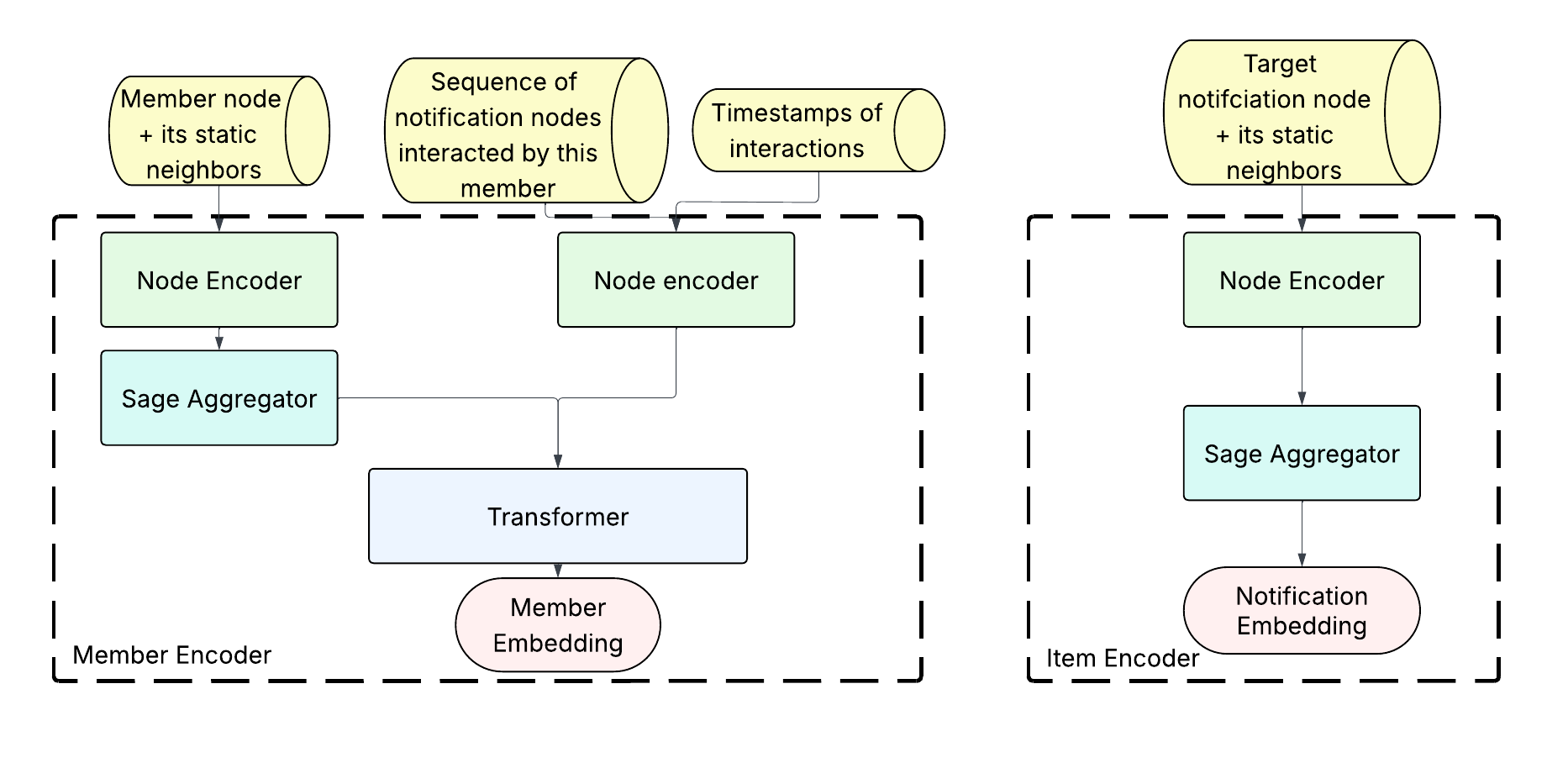}
\caption{High-level architecture of the member encoder (left) and the item encoder (right)}
\label{fig:gnn-arch}
\end{figure}
\begin{figure}[t]
\centering
\includegraphics[width=0.75\linewidth]{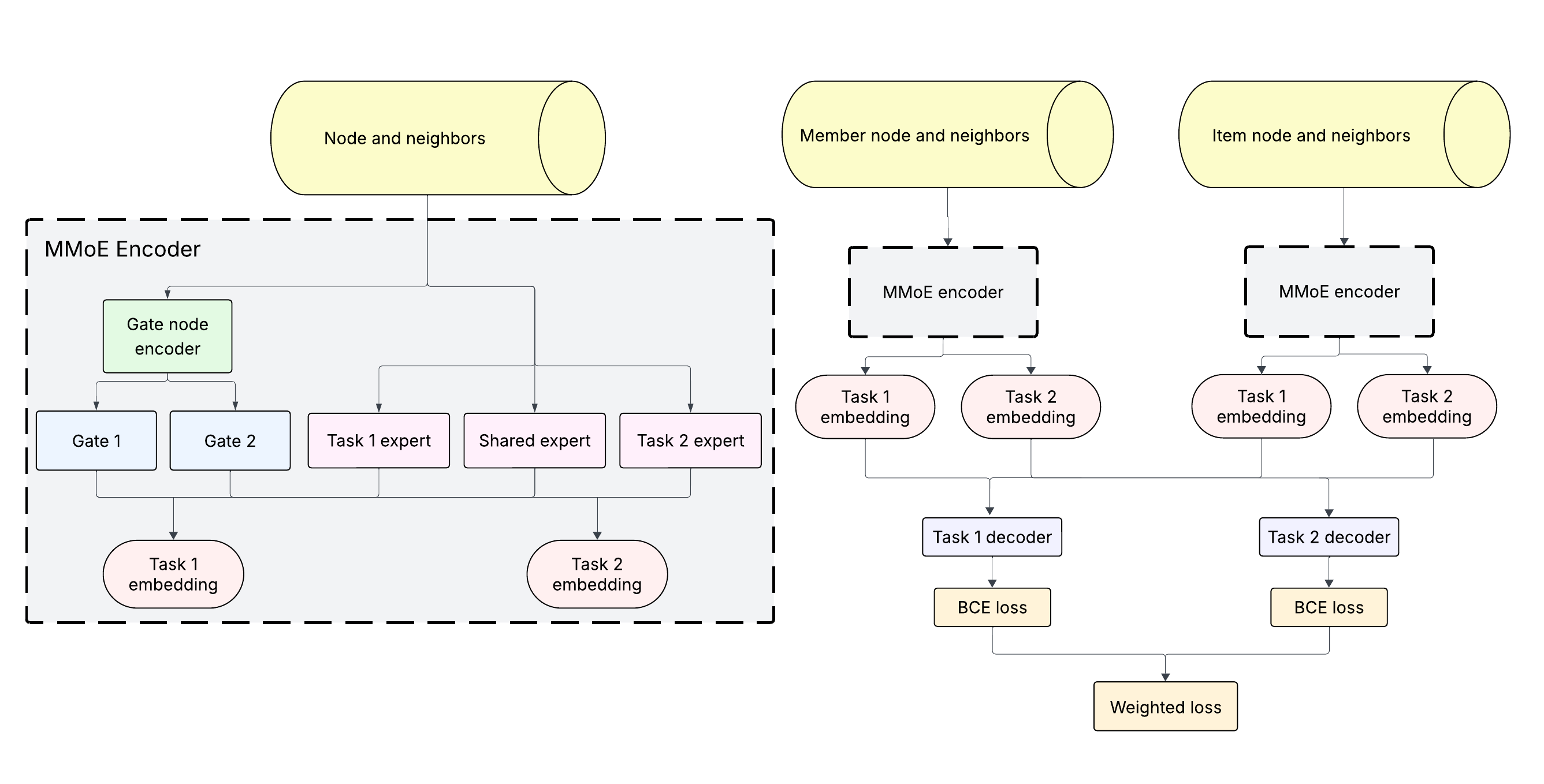}
\caption{High-level architecture of the Cross-domain GNN, featuring dual-tower entity encoders, multi-layer message passing, temporal modeling, and multi-task learning (MTL) components.}
\label{fig:gnn-arch}
\end{figure}

\subsection{Heterogeneous Message Passing Module}

We extend the GraphSAGE~\cite{hamilton2017inductive} and R-GCN~\cite{schlichtkrull2018modeling} frameworks to support multiple node and edge types in our heterogeneous graph. The message passing process is defined as:
\begin{equation}
\mathbf{m}_{u \to v}^{(l)} = f \left( \mathbf{h}_u^{(l-1)}, \mathbf{h}_v^{(l-1)}, t_{(u,v)} \right),
\end{equation}
\begin{equation}
\mathbf{h}_v^{(l)} = \sigma\Bigl( \mathrm{AGG}\bigl\{\mathbf{m}_{u \to v}^{(l)} : u\in \mathcal{N}(v)\bigr\} \Bigr),
\end{equation}

where:
\begin{itemize}
\item $\mathbf{h}_{v}^{(l)}$ is the hidden representation of node $v$ at layer $l$.
\item $t_{(u,v)}$ is the timestamp of the edge when available.
\item $f$ is a message function that transforms neighbor representations based on relationship type.
\item $\mathcal{N}(v)$ is the neighbor of node $v$.
\item $\mathrm{AGG}$ is an aggregation function (mean, sum, or attention-based) that combines messages from neighboring nodes.
\item $\sigma$ is a nonlinear activation function.
\end{itemize}

\subsection{Temporal Aggregation Module}
\label{sec:model:temporal}

To capture the dynamic nature of LinkedIn data, especially in time-sensitive domains like notifications, we incorporate a temporal aggregation module. For each member node, we:

\begin{enumerate}
\item Sample the chronologically most recent notification neighbors connected to the member
\item Transform these neighbors' features along with their edge timestamps using a dedicated encoder
\item Concatenate the standard message-passing embeddings with these temporally-aware representations
\item Process the combined representation through a transformer encoder to capture sequential patterns
\end{enumerate}

For timestamp encoding, we utilize Time2Vec~\cite{kazemi2019time2vec}, which transforms raw timestamps into learnable periodic and linear components:

\begin{equation}
\text{Time2Vec}(t)_i =
\begin{cases}
\omega_i t + \phi_i, & \text{if}\ i = 0 \\
\sin(\omega_i t + \phi_i), & \text{if}\ i > 0
    
\end{cases}
\end{equation}

where $\omega_i$ and $\phi_i$ are learnable parameters. This encoding allows the model to capture both linear time progression and periodic patterns at different frequencies, providing a rich representation of temporal dynamics.

\subsection{Multi-task Learning}

To simultaneously predict multiple objectives (e.g., Notification in-app click  and "Like" action), we implement MTL based on the Multi-gate Mixture-of-Experts (MMoE)\cite{ma2018modeling} architecture. We further incorporate task-specific and shared experts as proposed in Progressive Layered Extraction (PLE)\cite{tang2020progressive} to explicitly segregate task-specific and shared signals.

For each task $k$, the model learns a task-specific hidden representation $\mathbf{h}^k_v$ of node $v$ by using a gating network $\mathbf{g}^k$ to compute a weighted sum of embeddings from $n$ experts (including both dedicated task-specific experts and shared experts). Each expert is an entity encoder $\mathbf{f}_i$:

\begin{equation}
\mathbf{h}^k_v = \sum_{i=1}^{n}{\mathbf{g}^k_{i}(v)}\mathbf{f}_{i}(v).
\end{equation}
The gating networks are implemented as linear transformations of the node embeddings:
\begin{equation}
\mathbf{g}^k(v) = \mathrm{softmax}(W_{gk}N(v)),
\end{equation}
where $N$ is the node encoder and $W_{gk} \in \mathbb{R}^{n\times d}$ is a trainable matrix, with $n$ experts and $d$ being the node embedding dimensionality.

These task-specific embeddings for each member-item pair ($\mathbf{h}^k_m$, $\mathbf{h}^k_i$) are then processed through a task-specific multi-layer perceptron (MLP) decoder head ($\mathrm{MLP}^k$) to obtain the final predictions and compute a binary cross-entropy loss $\mathcal{L}_k$.

\begin{equation}
\hat{\mathbf{y}}^k_{(m,i)} = \mathrm{MLP}^k (\mathbf{h}^k_m, \mathbf{h}^k_i).
\end{equation}
Our final loss is a weighted sum across all $k$ tasks:
\begin{equation}
\mathcal{L} = \sum_{i=1}^{k} \alpha_k \cdot \mathcal{L}_k.
\end{equation}
where $\alpha_k$ are task importance weights that can be tuned to prioritize certain tasks.

\section{Production Pipeline}
\label{sec:production}

This section details how we productionized the cross-domain GNN at LinkedIn to support large-scale, real-time notification personalization while maintaining system efficiency and scalability.

\begin{figure}[t]
\centering
\includegraphics[width=1.0\linewidth]{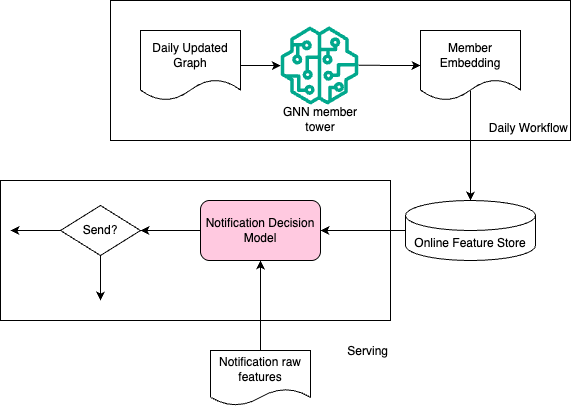}
\caption{Production pipeline for the cross-domain GNN: the system generates daily member embeddings, stores them in an online feature store, and feeds them to a downstream notification‑decision model that determines send eligibility and ranking.}
\label{fig:gnn-notif-pipeline}

\end{figure}

As illustrated in Figure~\ref{fig:gnn-notif-pipeline}, we decompose the Cross-domain GNN into two components: the \textit{member tower} and the \textit{item tower}, with only the member tower retained for daily inference to ensure serving-time efficiency and modularity.

Our production pipeline operates as follows:

\begin{enumerate}
\item \textbf{Daily Graph Construction:} We build an updated heterogeneous engagement graph that captures the latest member interactions across surfaces (notifications, email, feed).

\item \textbf{Member Embedding Generation:} The GNN member tower processes this graph to produce fresh member embeddings that encode behavioral patterns and content affinities.

\item \textbf{Feature Store Integration:} These embeddings are written to an online feature store, making them available for real-time serving.

\item \textbf{Decision Model Integration:} The notification decision model retrieves these embeddings along with other raw features to make personalized notification delivery decisions.

\end{enumerate}

\subsection{System Benefits}
\label{subsec
}

Our decoupled, asynchronous architecture provides several key advantages:

\begin{itemize}
\item \textbf{Daily Refresh:} Member embeddings are regenerated every day to reflect recent behavioral changes, ensuring representations remain current without requiring real-time GNN inference.

\item \textbf{Low-Latency Serving:} The decision model can retrieve precomputed embeddings from the feature store with minimal latency during notification delivery, meeting strict performance requirements.

\item \textbf{Cross-Surface Reusability:} The member embeddings can be repurposed across other LinkedIn surfaces, such as email ranking and feed recommendation, maximizing the return on computational investment.

\item \textbf{Modular Architecture for Optimized Performance:} By decoupling graph computation from real-time decision-making, we enable targeted optimization of each component. This separation allows the GNN to focus on producing high-quality representations, while the decision model is optimized for low-latency inference.

\end{itemize}

This architecture has enabled us to scale the cross-domain GNN to hundreds of millions of members while maintaining the performance characteristics required for a production recommendation system.

\section{Experiments and Results}
\label{sec:experiments}

This section presents our experimental methodology and results, focusing on the Notification use case as a concrete example of leveraging cross-domain signals to enhance model performance.

\subsection{Datasets and Experimental Setup}
\label{subsec:datasets_setup
}

\subsubsection{Cross-domain Graph Dataset}
\label{subsubsec:unigraph_datasets
}

To provide cross-domain signals for the GNN, we constructed a large-scale cross-domain Graph (15TB in disk space) incorporating data from multiple domains including Notification, Feed, and Email. The graph statistics are summarized in Tables~\ref{tab:biggraph-stats-node} and \ref{tab:biggraph-stats-edge}.

\begin{table}[ht]
\centering
\caption{Statistics of node types in the Cross-domain Graph}
\begin{tabular}{lcc}
\toprule
\textbf{Node Type} & \textbf{count} \\
\midrule
member & 1B \\
company & 20M \\
notification & 3B \\
post & 3B \\
email & 300M \\
\bottomrule
\end{tabular}

\label{tab:biggraph-stats-node}

\end{table}

\begin{table}[ht]
\centering
\caption{Statistics of edge types in the cross-domain Graph}
\begin{tabular}{lcc}
\toprule
\textbf{Edge type} & \textbf{count} \\
\midrule
notification-member-affinity & 44B \\
feed-member-affinity & 33B \\
member-click-post & 8B \\
member-click-notification & 918M \\
member-click-email & 300M \\
\bottomrule
\end{tabular}

\label{tab:biggraph-stats-edge}

\end{table}

The graph comprises:

\begin{itemize}
\item 8.6 billion nodes spanning members, content items, and organizations
\item Over 100 billion edges representing various interaction types
\item Diverse node features including content embeddings and categorical attributes
\end{itemize}

\subsubsection{Evaluation Datasets}
\label{subsubsec
}

We evaluate our approach using two key prediction tasks:

\paragraph{Notification Click-Through Rate (\textit{pClick})} A binary classification task that predicts whether a user will click on a presented notification. We partition the data into 80\% training, 10\% validation, and 10\% test splits, maintaining temporal consistency.

\begin{figure}[t]
\centering
\includegraphics[width=1.0\linewidth]{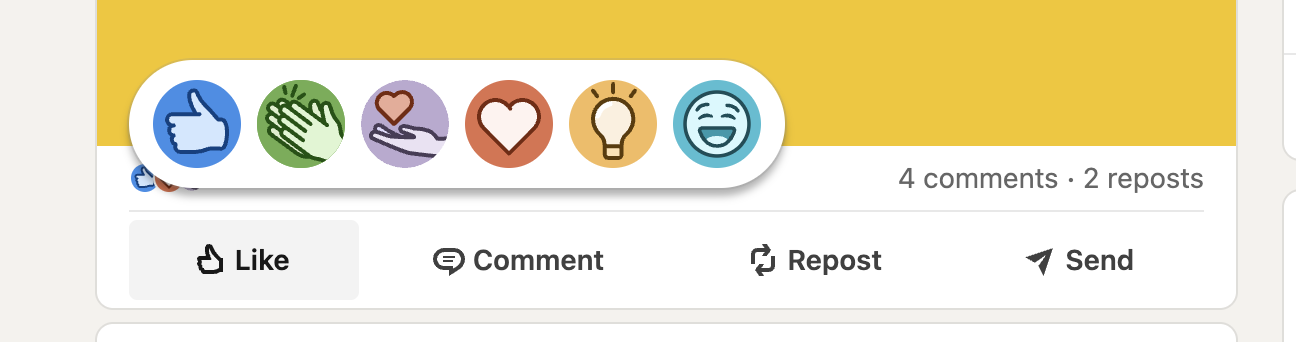}
\caption{Professional interactions that can occur after members click on notifications}
\label{fig:pi}
\end{figure}

\paragraph{Professional Interaction (\textit{pPI})} A derived metric that represents the likelihood that, after clicking on a notification, a member proceeds to perform meaningful downstream actions on the associated content (e.g., liking, commenting, extended dwell time). This metric captures deeper engagement beyond initial clicks, as illustrated in Figure~\ref{fig:pi}.

\subsection{Offline Evaluation}
Our offline experiments focus on three core innovations: cross-domain graph integration, temporal modeling, and MTL.

\subsubsection{Cross-Domain Graph Effectiveness}
\label{subsubsec
}

To measure the benefit of cross-domain signals, we compare the cross-domain GNN (trained on our unified heterogeneous graph spanning multiple domains) with a baseline GNN trained independently on domain-specific data. As shown in Table~\ref{tab:offline-graph}, the cross-domain GNN achieves an impressive AUC improvement of +8.089\%, demonstrating that shared signals between products significantly enhance predictive performance.

\begin{table}[ht]
\centering
\caption{Offline GNN AUC comparison from domain-specific and cross-domain graphs.}
\label{tab:offline-graph}
\begin{tabular}{lcc}
\toprule
\textbf{Graph} & \textbf{AUC lift}  \\
\midrule
Domain-specific     & --  \\
Cross-domain  & +8.089\% \\
\bottomrule
\end{tabular}
\end{table}

\subsubsection{Temporal Modeling Impact}
\label{subsubsec
}

We assess the contribution of temporal dynamics by enabling time-aware message passing in the cross-domain GNN. As shown in Table~\ref{tab:offline-mtl}, on the notification \textit{pClick} prediction task—a domain where user interests and content freshness rapidly evolve—we observe a substantial offline AUC lift of +0.265\% compared to the non-temporal counterpart. This highlights the importance of modeling edge recency and temporal patterns in real-time systems, particularly for time-sensitive content like notifications.

\subsubsection{Multi-Task Learning Benefits}
\label{subsubsec
}
Finally, we evaluate the effectiveness of applying a Multi-gate Mixture-of-Experts (MMoE) architecture to simultaneously predict multiple engagement signals. As shown in Table~\ref{tab:offline-mtl}, compared to single-task training, our multi-task cross-domain GNN improves AUC by +0.063\% for Notification \textit{pClick} and +0.327\% for Notification \textit{pPI}.

\begin{table}[ht]
\centering
\caption{Offline GNN AUC from Temporal Modeling and Multi-task Learning on Notification Click-Through Rate (\textit{pClick}) and professional interaction (\textit{pPI}) prediction tasks. Each AUC lift shows comparison against the corresponding metric from a non-temporal single-task Sage model baseline.}
\label{tab:offline-mtl}
\begin{tabular}{lcc}
\toprule
\textbf{Model} & \textbf{Metric} & \textbf{AUC lift}  \\
\midrule
Sage &\-- & --  \\
Temporal &\textit{pClick} & +0.265\%  \\
Sage MTL &\textit{pClick} & +0.063\% \\
Sage MTL &\textit{pPI} & +0.327\% \\
\bottomrule
\end{tabular}

\end{table}
These results underscore the benefits of parameter sharing across related tasks while allowing task-specific specialization. The larger relative improvement for \textit{pPI} suggests that MTL is particularly beneficial for more complex objectives that require deeper understanding of user intent and content value.

\subsection{Online A/B Test Results}

To evaluate the real-world impact of cross-domain GNN embeddings, we conducted a large-scale A/B experiment on LinkedIn's notification system. The experiment allocated 6\% of notification traffic to the treatment group using the cross-domain GNN-powered member embeddings, while 8\% served as the control group using the existing notification decision model without GNN embeddings. The experimental design ensured statistical significance while minimizing business risk during the evaluation period.

The cross-domain GNN embeddings were integrated into the second-pass ranker (SPR), which determines the final eligibility and prioritization of notifications before delivery. This integration point is critical as it allows the model to leverage rich contextual member representations at the decision boundary for notification sends.

\subsubsection{Experimental Setup}

For the experiment, we carefully controlled for potential confounding variables including:

\begin{itemize}
    \item \textbf{Member segments}: Ensuring balanced representation across engagement levels (highly active, moderately active, and dormant members)
    \item \textbf{Notification types}: Maintaining consistent distribution of content-based and connection-based notifications
    \item \textbf{Device types}: Balanced allocation across mobile and desktop users
    \item \textbf{Geographic regions}: Representative distribution across major markets
\end{itemize}

The experiment ran for a full 14-day period to account for weekly engagement patterns and ensure robust measurement of both immediate (CTR) and sustained (WAU) metrics.

\subsubsection{Metrics Framework}

We evaluated performance using a comprehensive set of metrics that capture both immediate engagement and longer-term retention:

\begin{itemize}
    \item \textbf{Sessions}: Defined as a contiguous period of member activity on LinkedIn, separated by at least 30 minutes of inactivity or a change in calendar day (PST). This metric primarily reflects the frequency of platform visits.
    
    \item \textbf{Weekly Active Users (WAU)}: Unique members who visit LinkedIn at least once within a rolling seven-day window. WAU is a core metric for assessing long-term retention and platform stickiness.
    
    \item \textbf{Click-Through Rate (CTR)}: Measured separately for in-app notifications and push notifications, quantifying the immediate relevance of delivered notifications.
    
    \item \textbf{Professional Interaction Rate (PIR)}: The proportion of clicked notifications that led to meaningful downstream engagement (commenting, liking, sharing, or extended dwell time). This metric captures the quality of engagement beyond initial clicks.
\end{itemize}

\subsubsection{Results Analysis}

Table~\ref{tab:online_results} presents the primary results from our online A/B test, showing consistent improvements across all key metrics.

\begin{table}[ht]
\centering
\caption{Online A/B test results comparing cross-domain GNN with baseline notification model. All improvements are statistically significant ($p < 0.01$).}
\label{tab:online_results}
\begin{tabular}{l|c}
\toprule
\textbf{Metric} & \textbf{Gain}  \\
\midrule
\textbf{Sessions} & +0.07\% \\
\textbf{WAU} & +0.10\% \\
\textbf{CTR (In-App)} & +0.62\% \\
\textbf{CTR (Push)} & +0.30\% \\
\bottomrule
\end{tabular}

\end{table}

The observed lifts in \textbf{Sessions} (+0.07\%) and \textbf{WAU} (+0.10\%) indicate that GNN-powered embeddings enhance member engagement across different activity levels. While modest in percentage terms, these gains are significant at LinkedIn’s scale.

The increase in session count suggests improved engagement among frequent users, likely driven by more personalized and timely notifications. The WAU gain points to better reactivation of less active members, highlighting the model’s ability to surface relevant content even with limited user history.

We also saw notable CTR improvements in both in-app (+0.62\%) and push (+0.30\%) channels. The stronger in-app CTR gain suggests higher responsiveness in work-focused contexts, while the improvement in push—despite its lower baseline—demonstrates the model’s robustness across delivery modalities.

\subsubsection{Segment-Level Performance}

To better understand the cross-domain GNN's impact across different member segments, we conducted a stratified analysis by engagement frequency, with the conclusion below:

\begin{itemize}
    \item \textbf{Highly active members}: We observed strong 0.10\% Sessions Gain for these members.
    
    \item \textbf{Moderately active members}: Largest overall gains across all metrics (Sessions: 0.36\%, WAU: +0.23\%, 1.16\% CTR), indicating that cross-domain signals provide particularly valuable information for this middle segment.
    
\end{itemize}

\subsubsection{Notification Type Analysis}

A breakdown of performance by notification type revealed differential impacts across content categories:

\begin{itemize}
    \item \textbf{Content-oriented notifications} (e.g., news, trending posts): Showed a CTR improvement of +0.93\%, benefiting from the GNN's ability to capture nuanced topical interests derived from member interactions across feed and email surfaces.
    
    \item \textbf{Connection-driven notifications} (e.g., job changes, birthdays, new connections): Exhibited a CTR increase of +2.62\%, enhanced by the GNN's strength in modeling relational proximity and behavioral affinity within the professional graph.
    
\end{itemize}

These results confirm that the Cross-domain GNN successfully captures and integrates multiple signal types—content affinity, social proximity, and temporal dynamics—to deliver more relevant notifications across the full spectrum of LinkedIn's notification ecosystem.

\section{Discussion}
\label{sec:discussion}

\subsection{Advantages of a Cross-domain GNN Approach}

Our empirical results show that unifying data from multiple domains (e.g., Feeds, Notifications, Jobs) into a single graph yields substantial gains for downstream prediction tasks. The +8.089\% AUC lift over domain-specific models demonstrates the value of cross-domain signals. This supports and extends previous findings in cross-domain recommendation to the graph setting~\cite{kang2019recommendation}.

Our architecture also provides two notable system-level benefits:

\begin{itemize}
    \item \textbf{Engineering Efficiency}: The unified pipeline removes redundant processing across domains, consolidating graph construction, feature extraction, and embedding generation. Multi-task learning further reduces maintenance overhead by avoiding per-domain models.
    
    \item \textbf{Representation Consistency}: Training on a comprehensive graph enables consistent node embeddings that capture behavioral patterns across LinkedIn, supporting more coherent user experiences.
\end{itemize}

The successful deployment of this system at LinkedIn scale under real-time constraints affirms the practicality of unified GNNs for industrial recommendation systems.

\subsection{Challenges and Future Work}

Scaling to billions of edges still poses challenges in terms of training efficiency and experimentation velocity. Our implementation requires approximately 32 GPU-days per training cycle. Additionally, balancing long-term versus short-term temporal signals remains an open area, as does tuning for heterogeneous domains with differing sparsity and noise profiles.

Looking forward, we plan to incorporate self-supervised pre-training (e.g., masked edge prediction) to improve generalization and enable faster adaptation. We also aim to explore continuous-time GNNs~\cite{rossi2020temporal} for finer-grained modeling of time-sensitive behaviors.

\section{Conclusion}
\label{sec:conclusion}

We presented a Cross-domain Graph Neural Network approach that unifies multiple LinkedIn product domains into a single heterogeneous graph for improved representation learning. By integrating data from notifications, feed, email, and other surfaces, the model captures rich cross-domain signals that enhance downstream performance.

Our architecture incorporates temporal modeling and multi-task learning (MTL) via a Mixture-of-Experts framework, enabling effective prediction of multiple engagement signals. The system achieves significant improvements in both offline AUC (+8.089\%) and online business metrics (+0.62\% CTR, +0.10\% WAU).

In production, our scalable pipeline generates fresh daily embeddings and supports low-latency serving across hundreds of millions of members. This demonstrates the feasibility and value of cross-domain GNNs in large-scale industrial recommender systems.

\bibliographystyle{plain} 
\bibliography{gnn}  


\end{document}